\documentclass[10pt,twocolumn,letterpaper]{article}

\usepackage{cvpr} 
\usepackage{times}
\usepackage{epsfig}
\usepackage{graphicx}
\usepackage{amsmath}
\usepackage{amssymb}
\usepackage{subcaption}
 \usepackage{booktabs}

\usepackage[pagebackref=true,breaklinks=true,letterpaper=true,colorlinks,bookmarks=false]{hyperref}

\begin{document}

    \title{Accurate Quantization for Gait Representation Learning}

\author{%
  Senmao Tian, \quad Haoyu Gao$^1$, \quad Gangyi Hong$^2$, \quad Shuyun Wang$^3$, \\ 
  JingJie Wang$^4$, \quad Xin Yu$^3$, \quad Shunli Zhang$^4$\thanks{Corresponding author.}  
  \\
  \small{$^{1}$Georgia Institute of Technology}, \small{$^{2}$Tsinghua University}\\ 
  \small{$^{3}$The University of Queensland}, \quad \small{$^{4}$Beijing Jiaotong University} \\
}

\maketitle
\thispagestyle{empty}

\begin{abstract}
  Existing deep learning methods have made significant progress in gait representation learning. Quantization can facilitate the application of gait models as a model-agnostic general compression technique. Typically, appearance-based models binarize inputs into silhouette sequences. However, mainstream quantization methods prioritize minimizing task loss over quantization error, which is detrimental to gait representation learning with binarized inputs. To address this, we propose a differentiable soft quantizer, which better simulates the gradient of the round function during backpropagation. This enables the network to learn from subtle input perturbations. However, our theoretical analysis and empirical studies reveal that directly applying the soft quantizer can hinder network convergence. We addressed this issue by adopting a two-stage training strategy, introducing a soft quantizer during the fine-tuning phase. However, in the first stage of training, we observed a significant change in the output distribution of different samples in the feature space compared to the full-precision network. It is this change that led to a loss in performance. Based on this, we propose an Inter-class Distance-guided Calibration (IDC) strategy to preserve the relative distance between the embeddings of samples with different labels. Extensive experiments validate the effectiveness of our approach, demonstrating state-of-the-art accuracy across various settings and datasets. Code is available at \href{https://github.com/Sheldon04/QGait}{link}.
\end{abstract}

\section{Introduction}
\label{sec:intro}

In recent years, gait representation learning has emerged as a prominent field due to its potential in long-range analysis based on walking patterns. Unlike other human-centered analysis technologies, gait, as a form of locomotion, is difficult to intentionally disguise in a consistent manner~\cite{Lam2007HumanGR, Connor2018BiometricRB}. Additionally, gait representation learning is robust against common covariates such as attire, carrying items, and standing conditions~\cite{Guan2015OnRT, Choudhury2016ClothingAC}. Therefore, gait representation learning methods hold significant promise for remotely analyzing walking dynamics in uncontrolled environments, thus offering broad application prospects.


\begin{figure}[!t]
    \centering
    \includegraphics[width=8.0cm]{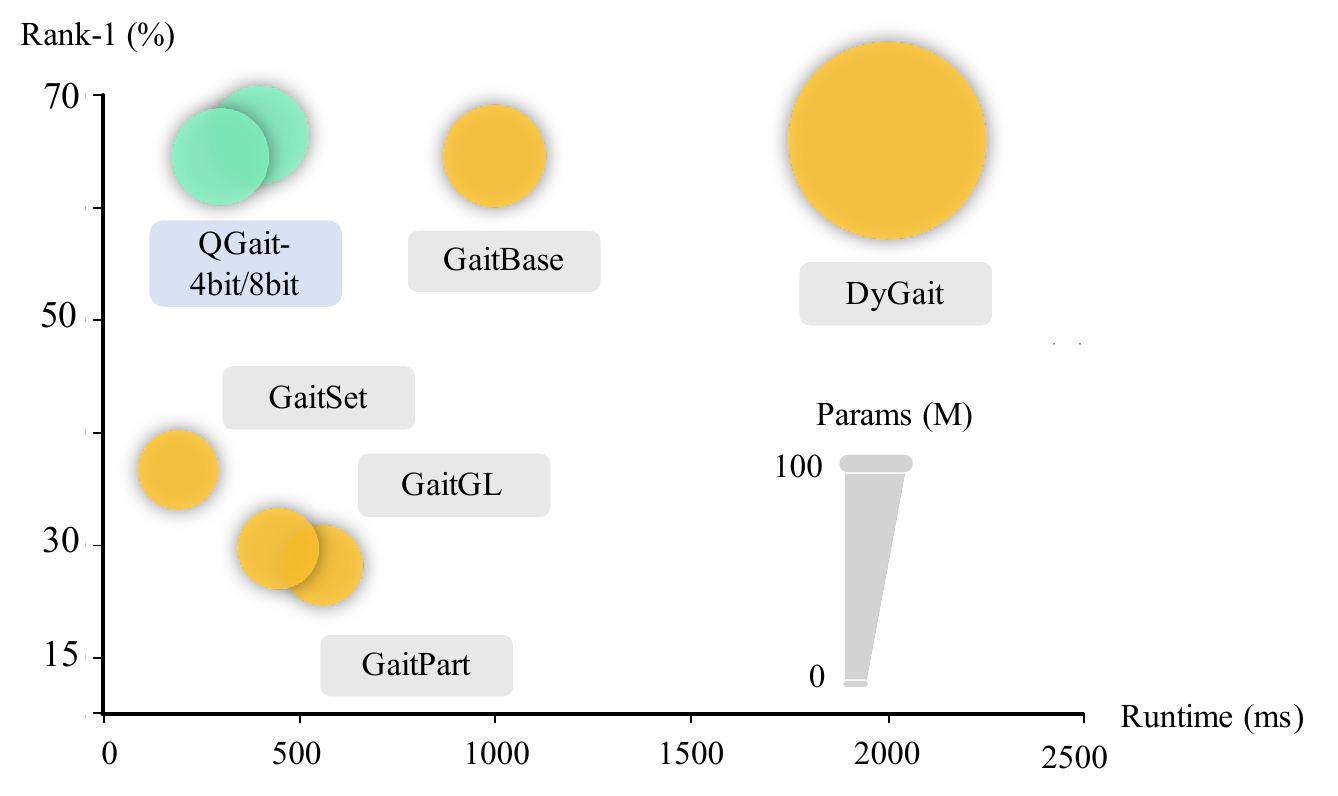}
    \caption{Parameters and performance comparison. QGait outperforms state-of-the-art methods with a relatively small number of parameters. Comparisons are performed on the Gait3D dataset~\cite{Zheng2022GaitRI}.}
    \label{fig:1}
\end{figure}

With the development of deep learning, gait representation learning methods based on deep features~\cite{Wu2017ACS, Makihara2017JointIA, Fan2020GaitPartTP, Hou2020GaitLN, Lin2020GaitRV, Lin2020GaitRW, guo2025gait, wang2025gaitadapt} have achieved notable results on various datasets~\cite{Yu2006AFF, Takemura2018MultiviewLP, Zheng2022GaitRI, Shen2022LidarGaitB3, Zhu2021GaitRI, Fan2022OpenGaitRG, Fan2023ExploringDM}. Currently, state-of-the-art gait representation learning technologies can achieve recognition distances exceeding hundreds of meters even with 4K high-definition cameras~\cite{Shen2022LidarGaitB3}. However, existing methods have scarcely considered the increasing computational resources required by the advanced edge devices. In real-world gait representation learning scenarios, especially when real-time processing of video streams is required, the inference overhead of gait models cannot be overlooked as Fig.~\ref{fig:real} shows. While methods based on CNNs~\cite{Fan2022OpenGaitRG, Zheng2022GaitRI} or Transformers~\cite{Fan2023ExploringDM} have demonstrated excellent performance, edge devices in deployment environments cannot meet the demands of these resource-intensive algorithms. This necessitates reducing the memory and computational burden of gait representation learning methods while preserving performance, enabling deployment on resource-constrained devices. Current efforts to compress and accelerate neural networks primarily focus on quantization~\cite{Zhou2016DoReFaNetTL, Choi2018PACTPC, Esser2019LearnedSS, Lee2021NetworkQW} and pruning~\cite{Lin2020HRankFP, Zhou2021LearningNM, He2018FilterPV}. Compared to pruning, quantization is more deployment-friendly and offers significant acceleration, particularly Quantization Aware Training~\cite{Jacob2017QuantizationAT}.


Existing quantization methods~\cite{Zhang2018LQNetsLQ, Baskin2018NICENI, Esser2019LearnedSS} have achieved decent results on some tasks, but applying them to gait representation learning tasks has failed to deliver comparable performance. These methods use a Straight-Through Estimator (STE)~\cite{Bengio2013EstimatingOP} to approximate the gradient of non-differentiable operations during the backward pass. While STE can smoothly transfer gradients, it is not sensitive to decimal-level changes. This may have no impact on other tasks but can be fatal for changes in gait posture. For example, in RGB input, the variation from 0 to 1 in a small area is not significant. However, for binarized gait silhouettes, this may indicate a change in posture. In this paper, we propose using parameterized soft quantization operators to prevent the network from accumulating quantization errors. Initially, we attempted to directly utilize a soft quantizer in the training process as a replacement for STE. However, through theoretical analysis and empirical validation, we have demonstrated that although this approach can simulate the quantization error introduced by non-differentiable operations, it significantly impacts the network's convergence. To harness the strengths of both STE and the soft quantizer while mitigating their respective limitations, we propose a two-stage training strategy. In the first stage, we employ STE to facilitate the network's convergence towards the vicinity of the optimal solution. Subsequently, we utilize the soft quantizer in the fine-tuning stage, enabling the simulation of decimal-level errors, which improves the network performance.

After quantization, the 8-bit models exhibit outstanding performance, surpassing even the full-precision model. However, when quantized to lower bit widths, the model's accuracy experiences significant degradation. Through analysis of the sample distribution in feature space, we observed significant changes in the relative positions of samples between the 4-bit quantized model and the full-precision model. This change could be a significant contributing factor to the observed decrease in accuracy. 
This insight prompts us to consider whether a calibration strategy could be employed to align the sample distribution of the low-bit models to high-bit models. A possible alignment solution is to distill the distribution from high-bit models. However, through an examination of the properties of naive distillation~\cite{Esser2019LearnedSS, Kim2019QKDQK, Yao2022ZeroQuantEA}, we found that such methods result in the low-bit model learning the numerical discrepancies between the outputs of the full-precision model and itself, which are discrepancies inherently introduced by quantization. Therefore, we propose an Inter-class Distance-guided Calibration strategy to focus more on the differences in sample distribution between classes rather than the numerical differences within classes.

Our contributions can be summarized as follows:
\begin{itemize}
\item We introduce a quantization-based gait representation network (QGait), which performs lossless compression and acceleration on existing methods.

\item We propose a two-stage quantization training strategy together with the IDC method, which simultaneously optimizes quantization error and task loss.

\item We conduct detailed experiments and the results demonstrate that our proposed QGait achieves comparable performance to full-precision networks with the lowest computational overhead.
\end{itemize}

\begin{figure*}[t!]
	\centering
	\includegraphics[width=0.85\linewidth]{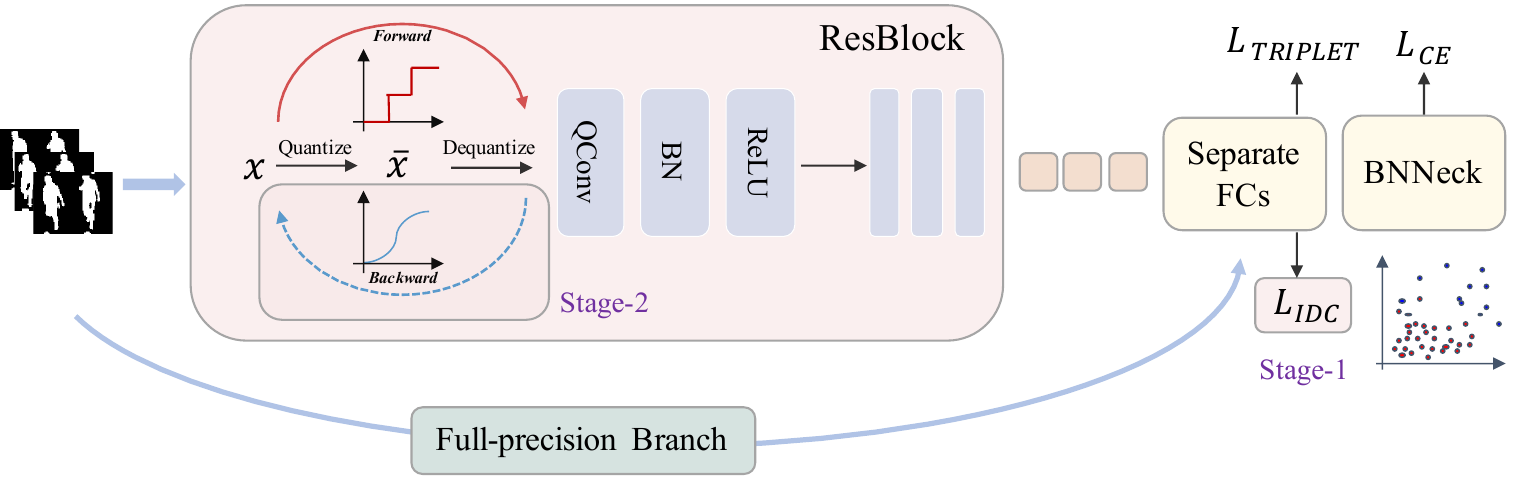} 
	\caption{The pipeline of our proposed QGait.}
	\label{fig:2}
\end{figure*}

\section{Related Work}
\subsection{Gait Representation Learning}
Gait representation learning methods can generally be classified into two categories: model-based methods~\cite{Li2020EndtoEndMG, Liao2017PoseBasedTN} and appearance-based methods~\cite{Wang2023DyGaitED, Shen2022GaitRW}.  Model-based approaches tend to utilize estimated human body models as input, such as 2D/3D poses and SMPL~\cite{Loper2023SMPLAS}.  For instance, GaitGraph~\cite{Teepe2021GaitgraphGC} employs graph convolutional networks for learning gait representations based on 2D skeletons, SMPLGait~\cite{Zheng2022GaitRI} leverages the 3D geometric information of SMPL models to enhance the learning of gait appearance features, and GaitTR~\cite{Zhang2022SpatialTN} combines transformer and convolutional layers to represent spatial and temporal information respectively.  However, although model-based methods theoretically exhibit robustness to factors like carrying items and clothing, they often perform poorly due to the error accumulation in the pre-processing stage while entailing high computational costs, thereby potentially lacking practicality in certain real-world scenarios. On the other hand, appearance-based methods directly learn shape features from input videos, offering simplicity and ease of use while preserving privacy.  With the rise of deep learning, most current appearance-based works focus on spatial feature extraction and gait temporal modeling.  Specifically, GaitSet~\cite{Chao2018GaitSetRG} treats gait sequences as a set for the first time, employing the maximum function to compress sequences of frame-level spatial features. GaitPart~\cite{Fan2020GaitPartTP} meticulously explores minor differences in various parts of the silhouette to model periodic gait features.  GaitGL~\cite{Lin2020GaitRV} addresses the limitations of spatially global gait representations in neglecting details, and locally based descriptors in failing to capture relationships between adjacent parts, by developing global and local convolutional layers.  Building upon these works, GaitBase~\cite{Fan2022OpenGaitRG} significantly simplifies gait modeling and achieves excellent results with its simple yet effective design. Considering that existing methods often lack a practical perspective, it becomes necessary to design a lightweight network to aid gait representation models in real-world applications.  To the best of our knowledge, the proposed QGait is the first quantization-based compressed gait representation network.

\begin{figure*}[!t]
	\centering
    \subfloat[]{
        \includegraphics[width=0.32\linewidth]
        {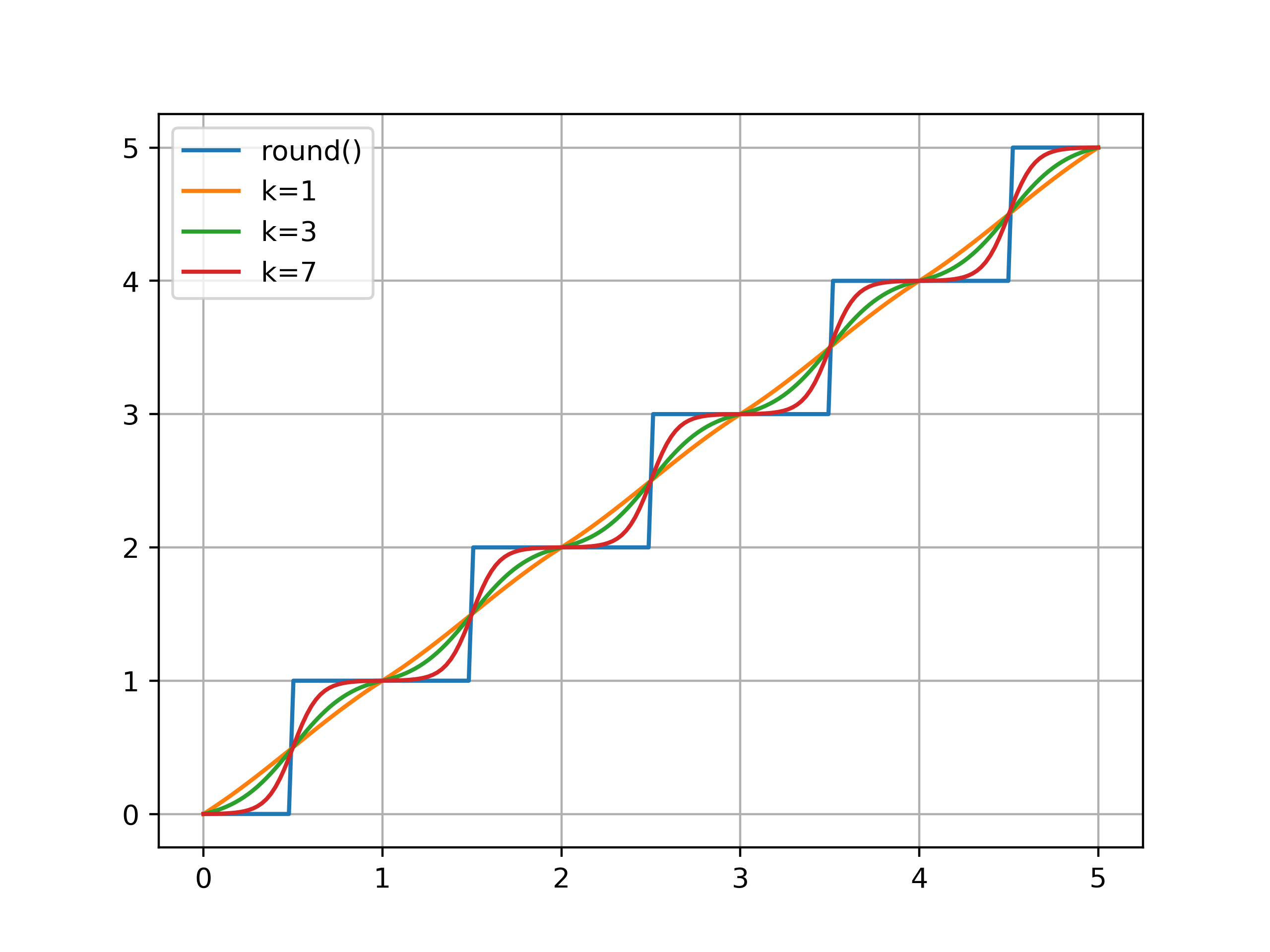}
    }
    \subfloat[]{
        \includegraphics[width=0.32\linewidth]
        {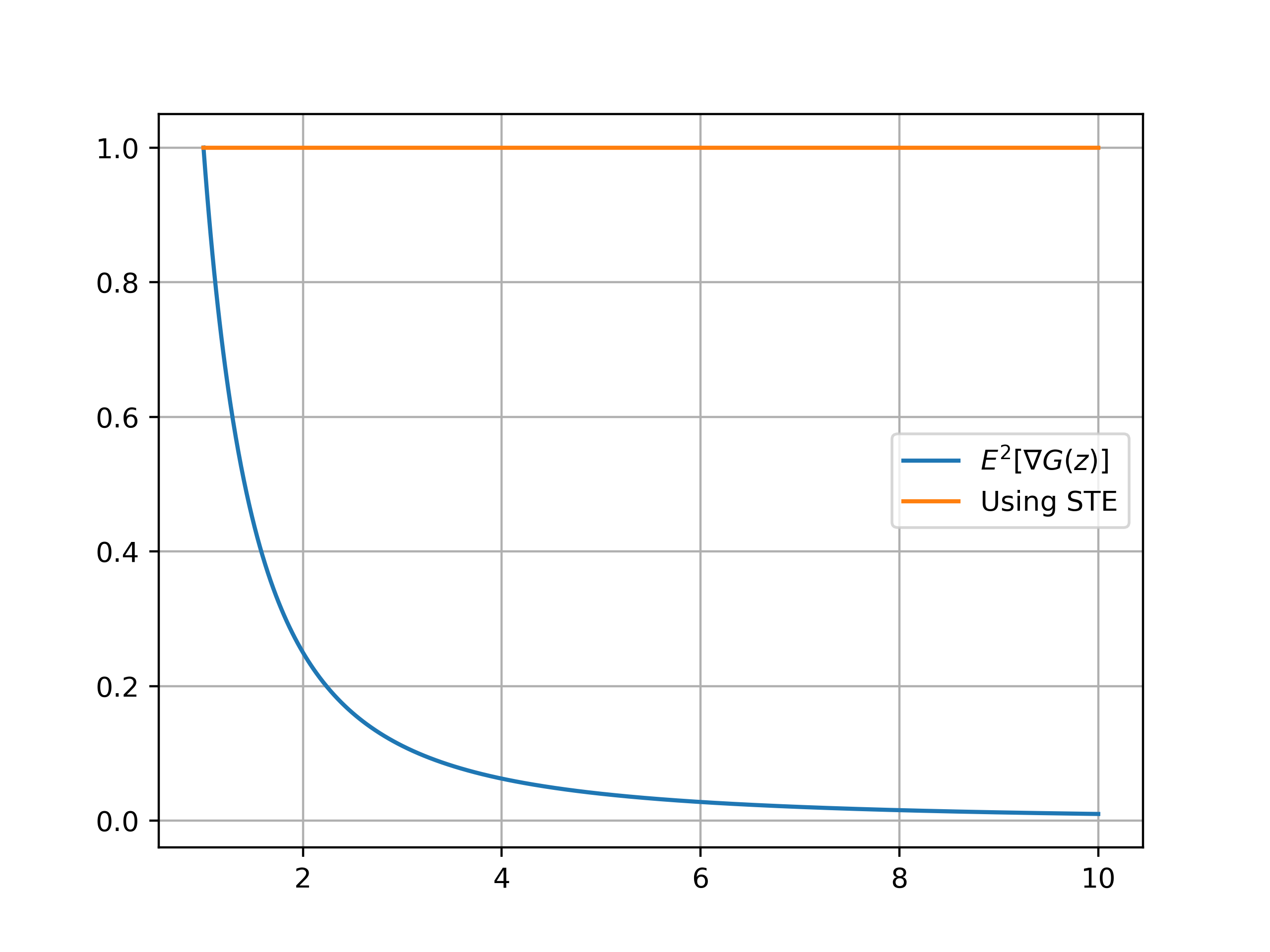}
    }
    \subfloat[]{
        \includegraphics[width=0.32\linewidth]
        {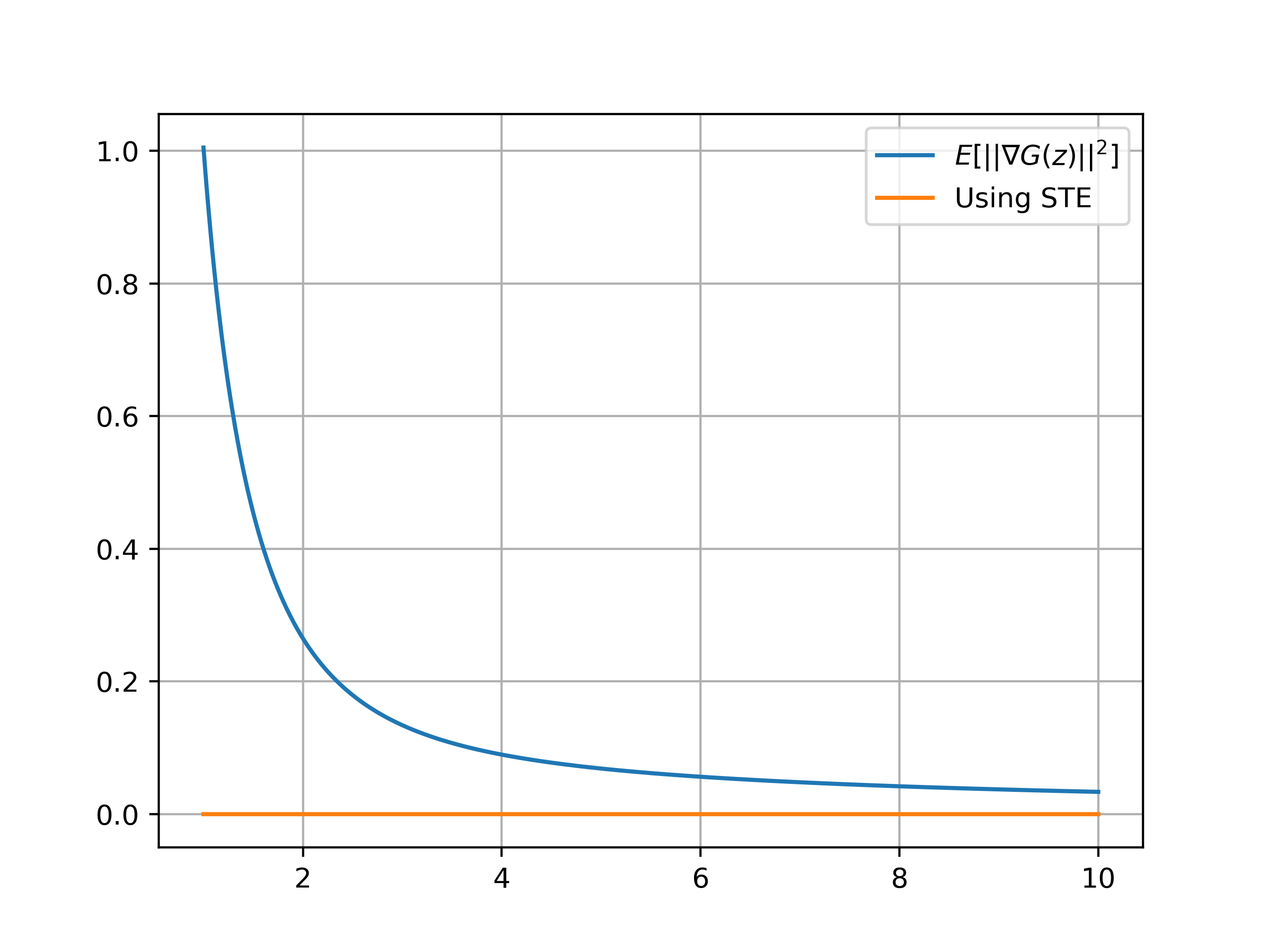}
    }
	\caption{(a) Visualization of the round function along with the graph of ${\theta}_{k}(x)$ used for its approximation when $k$ takes different values. (b) (c) Visualization of $\mathbb{E}^2\left[ \nabla G(z)\right]$ and $\mathbb{E}[\|\nabla G(z)\|^2] $.}
	\label{fig:3}
\end{figure*}

\subsection{Network Quantization}
Quantization~\cite{Chen2024AdaptiveQW, Cai2023BinarizedSC, Rastegari2016XNORNetIC} is a widely adopted technique in computer vision for compressing and accelerating neural networks. The advantage of quantization is that it can be easily combined with pruning, distillation, and other methods to further compress the network. Depending on the mapping strategy, quantization can be categorized into uniform and non-uniform~\cite{Liu2021NonuniformtoUniformQT} quantization. Non-uniform quantization necessitates specific hardware support, thus current research predominantly focuses on uniform quantization exploration. Uniform quantization methods encompass Post-Training Quantization (PTQ)~\cite{Shang2022PostTrainingQO} and Quantization Aware Training (QAT). PTQ has gained popularity for its ability to quantize models without requiring retraining. However, it suffers from accuracy degradation due to its heavy reliance on input data for setting quantization parameters, leading to poor alignment between parameters and weights. QAT addresses this issue by training with inserted fake quantization nodes~\cite{Jacob2017QuantizationAT}, allowing network weights to adapt to quantization errors and enhancing the accuracy of quantized models. Early works like Dorefa~\cite{Zhou2016DoReFaNetTL} achieve network acceleration by quantizing activation values, weights, and gradients simultaneously. Further advancements such as PACT~\cite{Choi2018PACTPC}, utilize learnable parameters to clip activation values to gain more reasonable quantization intervals. LSQ~\cite{Esser2019LearnedSS}, one of the most widely used quantization methods, introduces a novel approach to estimate and scale the task loss gradient on the quantizer step size for each weight and activation layer, enabling the quantization step size to be learned with network optimization and achieving commendable results across various tasks. Given the satisfactory accuracy and notable acceleration achieved by quantized networks, quantization-based model compression has been extensively studied in various domains~\cite{Tian2023CABMCB, Yao2022ZeroQuantEA, Lin2023AWQAW}, demonstrating the efficacy of quantization schemes. However, the impact of quantization in gait representation learning remains underexplored. In this paper, we introduce QGait, which, built upon optimized quantization training strategies and tailored to the characteristics of gait representation learning tasks, unveils the effectiveness of quantization in compressing gait representation networks for the first time, offering novel perspectives and insights into gait representation learning research.

\section{Method}
\subsection{Preliminaries}

\paragraph{Network Architecture} 
To strike a balance between universality and effectiveness, among numerous methods, we opt for GaitBase as the baseline model for compression. For appearance-based methods, given a set of captured RGB images $I = \{I^{i}|I^{i} \in \mathbb{R}^{T\times3\times H\times W}\}_{i=1}^{N}$, the input of the models are designed as binary silhouette sequences $S = \{S^{i}|S^{i} \in \mathbb{R}^{T\times1\times H\times W}\}_{i=1}^{N}$, where $T$ represents the number of frames in the sequence and $N$ is the number of data points. For GaitBase, the model $\mathcal{M}$ can be formulated as:
\begin{equation}
 \mathcal{X} = \mathcal{M}(S) = \mathcal{F}\circ\mathcal{P}\circ\mathcal{B}(S),   
\end{equation}
where $\mathcal{X}$ denotes the embeddings, $\mathcal{F}$ is the separate fully connected layer, $\mathcal{P}$ is the temporal pooling and horizontal pooling layer, $\mathcal{B}$ is the backbone network, and $\circ$ denotes the connection among network parts. Given the BNNecks module $\mathcal{K}$, $\mathcal{O} = \mathcal{K}(\mathcal{X})$ is the logits vector.

\paragraph{Quantization Framework} 
\begin{figure*}[t!]
	\centering
	\includegraphics[width=0.8\linewidth]{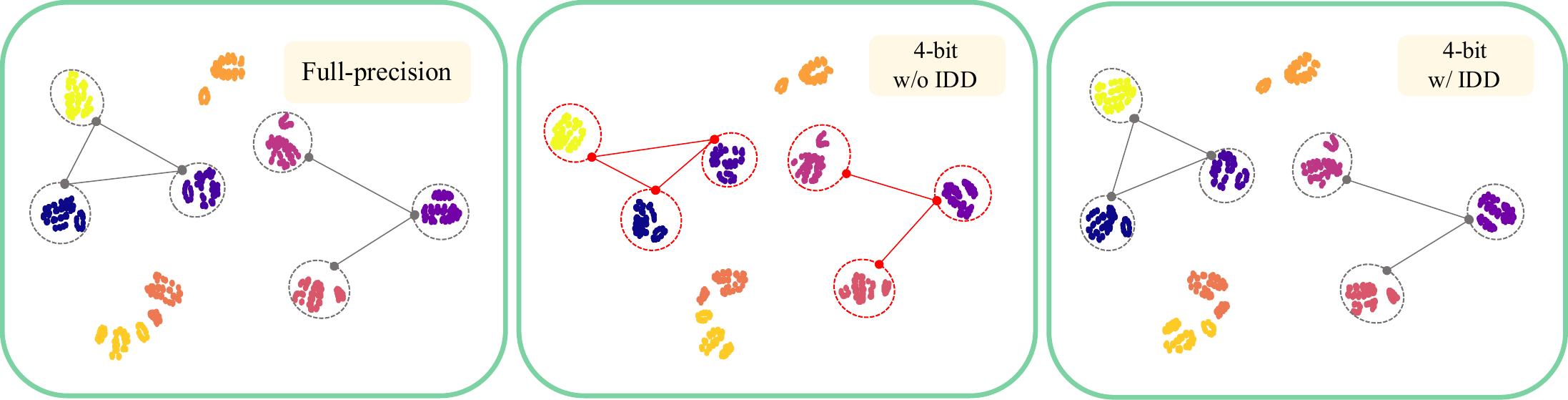} 
	\caption{Visualization using t-SNE of the full-precision network, 4-bit network without IDC and 4-bit network with IDC. The results indicate that IDC can significantly mitigate the changes in inter-class distances.}
	\label{fig:4}
\end{figure*}
For the quantized gait representation network, the weight and activation values of
computing units (such as convolutional and linear layers) are compressed to low bit-widths by the following point-wise quantization
functions:
\begin{equation}
    \bar{x}=\lfloor clamp(\frac{x}{v})\rceil,
\end{equation}
\begin{equation}
    \\\hat{x}=\bar{x} \cdot v,
\end{equation}
where $x$ denotes either weights or activations of a specific layer, $clamp(\frac{x}{v})$ returns $\frac{x}{v}$ with values below $r_1$ set to $r_1$ and values above $r_2$ set to $r_2$ ($r_1$ and $r_2$ are scopes set based on quantization type and bit-width), $v$ is a learnable parameter that adjusts the quantization step size and $\lfloor * \rceil$ is the round function. For example, given a quantization bit-width $b$, $r_1$ and $r_2$ is set to $0$ and $2^b-1$ respectively for unsigned input and $-2^{b-1}$ and $2^{b-1}-1$ for signed input. Since the round function is non-differentiable, the derivative of $\hat{x}$ with respect to $\bar{x}$ during the backward pass~\cite{Esser2019LearnedSS, Choi2018PACTPC, Zhou2016DoReFaNetTL} can be represented as:
\begin{equation}
    \frac{\partial \hat{x}}{\partial x}=\left\{\begin{array}{ll}
    1, & \text {if} \quad r_1 < \frac{x}{v} < r_2, \\
    0, & \text { otherwise. }
    \end{array}\right.
\end{equation}

\paragraph{Overall Objective} 
For QAT, global optimization learning algorithms optimize a suitable objective function $\mathcal{L}$ by observing the training data as a whole to learn the parameters of deep networks:
\begin{equation}
    \theta=\arg\min_\theta\mathcal{L}_{joint}(\theta;S,Y),
\end{equation}
where $Y$ denotes the corresponding labels and $\mathcal{L}_{joint}$ represents the implicit combination of task loss $\mathcal{L_T}$ and quantization error $\mathcal{L_Q}$. The pipeline of QGait is shown in Fig.~\ref{fig:2}.

\subsection{Training with Soft Quantizer}

STE serves as a classical method for estimating gradients, proven to be simple yet effective in many tasks. For some learning-based quantization methods, using STE to estimate quantization errors has yielded promising results. This is because, during quantization-aware training, algorithms tend to minimize task loss $\mathcal{L_T}$ rather than quantization error $\mathcal{L_Q}$~\cite{Esser2019LearnedSS, Liu2021NonuniformtoUniformQT}. However, in the context of gait representation learning tasks, particularly in appearance-based approaches with binarized input, even minor numerical perturbations can lead to vastly different features. Therefore, it should be helpful to use a differentiable function to approximate the step function to enhance the performance of gait representation learning tasks. For the round function, the maximum quantization error occurs at the midpoint between two adjacent approximate values. Additionally, we hope that the derivative of this function varies periodically with changes in input values. A function that meets the above requirements can be defined as follows:
\begin{equation}
    {\theta}_{k}(x)=\lfloor x\rfloor+\frac{1}{2} \frac{\tanh (k d)}{\tanh (k / 2)}+\frac{1}{2}, \quad k \geq 1,
\end{equation}
where $d = x - \lfloor x\rfloor - 0.5$ is the range of error, $k$ is set to the magnitude of gradient changes, and $\lfloor *\rfloor$ denotes the floor function. The derivative of $\hat{x}$ with respect to $\bar{x}$ during the backward pass becomes:
\begin{equation}
    \frac{\partial \hat{x}}{\partial x}=\left\{\begin{array}{ll}
    \frac{\partial {\theta}_{k}(x)}{\partial x}, & \text { if } \quad r_1 < \frac{x}{v} < r_2, \\
    0, & \text { otherwise. }
    \end{array}\right.
\end{equation}

However, from an optimization perspective, while such a function can simulate errors effectively by setting different values of $k$, it complicates the convergence of the network. Let $G(x)$ be a convex function that is L-smooth. Generally, our iterations are given by $x_{t+1} = x_t - \eta v_t$, where $\eta$ is the step size and $v_t$ is a random direction. Then we have:
\begin{equation}
\label{eq7}
\begin{aligned}
    & \mathbb{E}\left[G(x_{t+1})-G(x_t)\right] \\
    & \leq\mathbb{E}\left[\langle x_{t+1}-x_t,\nabla G(x_t)\rangle+\frac{L}{2}\|x_{t+1}-x_t\|^2\right] \\
    &=-\eta\left\langle\mathbb{E}(v_t),\nabla G(x_t)\right\rangle+\frac{\eta^2L}2\mathbb{E}[\|v_t\|^2],
\end{aligned}    
\end{equation}
where $L$ is the smoothness coefficient. Under the framework of stochastic gradient descent (SGD), we can approximately consider $v_t = \nabla G(x_t)$.

Next, let's analyze the two terms on the right-hand side of Eq.~\ref{eq7}. For the first term, the gradient is always set to $1$ during the backward pass when using STE. In most machine learning libraries, the gradient of floor functions is set to $0$. So we only need to consider part of the ${\theta}_{k}(x)$: 
\begin{equation}
    G(\mathbf{z}) =\frac{\tanh (\mathbf{z})}{2 \tanh (k / 2)}, \quad \mathbf{z} \in [-0.5k, 0.5k).
\end{equation}
We can then compute the gradient of $G(x)$ as:
\begin{equation}
     \nabla G(\mathbf{z}) =  \frac{1}{2} \operatorname{coth}\left(\frac{k}{2}\right) \operatorname{sech}^{2}(\mathbf{z}).
\end{equation}
Assuming the random variable $\mathbf{z}$ follows a uniform distribution, we can compute the first term as:
\begin{equation}
\label{eq10}
\begin{aligned}
    \mathbb{E}^2\left[ \nabla G(\mathbf{z})\right] 
    &= {\left(\int_{-\frac{k}{2}}^{\frac{k}{2}} \frac{1}{2} \operatorname{coth}\left(\frac{k}{2}\right)(\operatorname{sech}(\mathbf{z}))^{2} \frac{1}{k} d\mathbf{z}\right)}^2 \\
    &= \frac{1}{k^2} < 1,
\end{aligned}  
\end{equation}
which is a function involving $k$ that is consistently less than $1$ as Fig.~\ref{fig:3}(b) shows. As $k$ increases, this term tends to approach $0$, indicating that the optimization process will become increasingly challenging to continue. However, we aim for as much descent as possible at each iteration, so STE is better for the first term. As for the second term, we can compute that:
\begin{equation}
\label{eq11}
\begin{aligned}
    \mathbb{E}[\|\nabla G(\mathbf{z})\|^2] 
    &= \int_{-\frac k2}^{\frac k2}\left(\frac12\coth\left(\frac k2\right)(\mathrm{sech}(\mathbf{z}))^2\right)^2\frac1kd\mathbf{z}\\
    &= \frac{\cosh(k)+2}{3 k \sinh{k}} > 0,
\end{aligned}  
\end{equation}
which is consistently greater than $0$. Combining Eq.~\ref{eq7}, Eq.~\ref{eq10}, and Eq.~\ref{eq11}, we can see that while the soft quantizer simulates quantization error, STE is superior in terms of convergence as Fig.~\ref{fig:3}(c) shows. Therefore, during the initial training phase, STE is utilized to minimize task loss, while during the second fine-tuning phase, the soft quantizer is employed to balance quantization error and task loss.

\subsection{Inter-class Distance-guided Calibration}

\begin{table*}[!t]
  \centering
     \caption{Results on Gait3D dataset. 'w' and 'a' represent the bit of the weight and activation respectively.}
    \begin{tabular}{c|ccccccc}
    \toprule[1.2pt]
    Methods & Bit-width & BitOPs (G) & Rank-1 (\%) & Rank-5 & Rank-10& mAP & mINP \\
    \midrule[0.6pt]
    GaitSet~\cite{Chao2018GaitSetRG} & 32    & 10.68  & 36.75  & 58.66  & 64.20  & 30.21  & 17.55 \\
    GaitPart~\cite{Fan2020GaitPartTP} & 32    & 10.67  & 28.44  & 47.58  & 53.70  & 22.01  & 12.32 \\
    GaitGL~\cite{Lin2020GaitRV} & 32    & 41.64  & 29.88  & 48.69  & 54.42  & 22.30  & 13.50 \\
    DyGait~\cite{Wang2023DyGaitED} & 32    & 652.22  & 66.30  & 80.80  & 86.10  & 56.40  & 37.30 \\
    GaitBase~\cite{Fan2022OpenGaitRG} & 32    & 118.30  & 64.40  & 81.50  & 85.80  & 55.28  & 36.73  \\
    \midrule[0.6pt]
    Dorefa-4~\cite{Zhou2016DoReFaNetTL} & w4/a4  & -  & 60.00  & 78.30  & 83.20  & 50.32  & 32.66 \\
    PACT-4~\cite{Choi2018PACTPC} & w4/a4  & -  & 59.80  & 78.00  & 82.80  & 49.56  & 31.53 \\
    LSQ-4~\cite{Esser2019LearnedSS} & w4/a4  & -  & 63.80  & 79.60  & \textbf{86.30}  & 54.22  & 35.71 \\
    
    QGait-4 & w4/a4  & 1.85  & \textbf{64.60}  & \textbf{81.20}  & 86.20  & \textbf{54.29}  & \textbf{35.76} \\
    \midrule[0.6pt]
    Dorefa-8~\cite{Zhou2016DoReFaNetTL} & w8/a8  & -  & 64.30  & 80.80  & 86.10  & 54.34  & 35.87 \\
    PACT-8~\cite{Choi2018PACTPC} & w8/a8  & -  & 63.30  & 80.50  & 85.20  & 53.41  & 34.64 \\
    LSQ-8~\cite{Esser2019LearnedSS} & w8/a8  & -  & 64.80  & \textbf{81.40}  & 85.90  & 55.15  & 36.76 \\
    QGait-8 & w8/a8  & 7.39  & \textbf{66.50}  & 81.30  & \textbf{86.20}  & \textbf{55.82}  & \textbf{37.05} \\
    \bottomrule[1.2pt]
    \end{tabular}
   \label{tab:1}%
\end{table*}%

In the preceding sections, we mitigated the impact of quantization error on prediction results by introducing the soft quantizer. However, on some complex datasets~\cite{Takemura2018MultiviewLP, Zheng2022GaitRI}, transitioning from 32-bit to 8-bit quantization still inevitably incurs some loss due to inherent information loss in the process of data discretization. Moreover, we observed that for the embeddings outputted by the 4-bit network, the relative distance between samples of different classes has changed. This inspires us to explore whether we can leverage information from high-bit networks to guide the optimization process of low-bit networks.

Given an input $S$, the output embedding vector is $\mathcal{X} = \mathcal{M}(S)$. For $\mathcal{M}$ with high bit $\mathcal{M}_H$ and low bit $\mathcal{M}_L$, the logits vectors are denoted as $\mathcal{O}_H$ and $\mathcal{O}_L$ respectively. Naive Knowledge Distillation (KD) approaches measure the distribution difference between two vectors. The probability vectors are calculated as:
\begin{equation}
q(\mathcal{O}^{(i)})=\frac{\exp(\mathcal{O}^{(i)}/\mathcal{T})}{\sum_{m}\exp(\mathcal{O}^{(m)}/\mathcal{T})},
\end{equation}
where $i$ and $m$ are class indices, and $\mathcal{T}$ is the temperature. Generally, the objective is realized by minimizing KL divergence:
\begin{equation*}
\mathcal{L}_{\mathrm{KL}}\left(q(\mathcal{O}_H)||q(\mathcal{O}_L)\right)=\sum_{i}q(\mathcal{O}_H)^{(i)}\log\left(\frac{q(\mathcal{O}_H)^{(i)}}{q(\mathcal{O}_L)^{(i)}}\right).
\end{equation*}
For the well-trained low-bit model, we assume that the KL divergence loss is minimized, aligning the predicted probability density with that of the high-bit model. For any pair of class indices $i, j$, it gives:
\begin{equation*}
    \frac{\exp\left[\mathcal{O}_H^{(i)}/\mathcal{T}\right]}{\exp\left[(\mathcal{O}_H^{(j)})/\mathcal{T}\right]}=\frac{\exp\left[(\mathcal{O}_L^{(i)})/\mathcal{T}\right]}{\exp\left[(\mathcal{O}_L^{(j)})/\mathcal{T}\right]},
\end{equation*}
If we simplify the above equation and generalize it to all classes, we obtain:$\frac{\sigma(\mathcal{O}_H)^2}{\sigma(\mathcal{O}_L)^2}=1,$ where $\sigma$ is the function of standard deviation. In most practical applications, discrete quantization tends to reduce the variance of the data, as the quantization process essentially reduces the precision of the data, causing values that may be very close to each other to be merged into a single value or range. This merging reduces the differences between data points, thus typically leading to a decrease in variance. The experiments also confirm this, showing that the variance of logits decreases from $0.1033$ to $0.05273$ after quantization. Therefore, the aforementioned optimization process essentially simulates the numerical disparities brought about by quantization. 

Inspired by the remarkable performance of triplet loss in gait representation learning~\cite{Schroff2015FaceNetAU, Fan2022OpenGaitRG}, we visualize the distribution of embedding vectors from different samples as shown in Fig.~\ref{fig:4}. Both low-bit and high-bit networks can effectively cluster samples of the same class. Despite minor differences in inner-class distributions, the impact of quantization-induced numerical disparities on performance is negligible compared to noticeable changes in inter-class distances (dashed lines in Fig.~\ref{fig:4}). Based on the observation of heterogeneous distances, we decided to approach a calibration from the perspective of distance distribution. Considering a mini-batch of data, the probability vectors guided by the self-similarity of distances between samples' embeddings are defined as:
\begin{equation}
\begin{aligned}
q^*(\mathcal{X}^{(r)},\mathcal{X}^{(s)})&=\frac{\exp(-d(\mathcal{X}^{(r)}, \mathcal{X}^{(s)}))}{\sum_{u}\exp(-d(\mathcal{X}^{(r)}, \mathcal{X}^{(u)}))}, 
\\ &\text{s.t.}\quad Y_r \neq Y_s, Y_r \neq Y_u, 
\end{aligned}
\end{equation}
where $Y_*$ denotes the labels of samples, $d$ indicates the Euclidean distance, $r, s$ and $u$ are the sample indices. The final objective function can be formulated as:
\begin{equation}
\begin{aligned}
&\mathcal{L}_{\mathrm{IDC}}\left(q^*(\mathcal{X}_H^{(r)},\mathcal{X}_H^{(s)})||q^*(\mathcal{X}_L^{(r)},\mathcal{X}_L^{(s)})\right) \\
&=\sum_{r,s}q^*(\mathcal{X}_H^{(r)},\mathcal{X}_H^{(s)})^{(r,s)}\log\left(\frac{q^*(\mathcal{X}_H^{(r)},\mathcal{X}_H^{(s)})^{(r,s)}}{q^*(\mathcal{X}_L^{(r)},\mathcal{X}_L^{(s)})^{(r,s)}}\right).
\end{aligned}
\end{equation}

\section{Experimental Results}
To validate the generalization and effectiveness of our approach, we performed comprehensive experiments on the mainstream datasets and compared our approach with state-of-the-art gait methods and quantization methods. We selected representative methods Dorefa~\cite{Zhou2016DoReFaNetTL}, PACT~\cite{Choi2018PACTPC}, and LSQ~\cite{Esser2019LearnedSS} as the baselines for quantization. QGait models in subsection \ref{sec:4.2} are implemented based on GaitBase~\cite{Fan2022OpenGaitRG}. Other quantized models are reported in subsection \ref{sec:4.4}. 
The code implementation is based on PyTorch 2.0, and all experiments were conducted using RTX A6000s.

\begin{table}[!t]
  \centering
        \caption{Results on GREW dataset.}
    \begin{tabular}{c|cccc}
    \toprule[1.2pt]
    Methods & Bit & BitOPs & Rank-1 & Rank-5  \\
    \midrule[0.6pt]
    GaitSet  & 32    & 50.68  & 46.30 & 63.62  \\
    GaitPart& 32    & 31.68  & 43.99  & 60.74\\
    GaitGL  & 32    & 234.21  & 47.28 & 63.88 \\
    DyGait  & 32    &  867.14  & 71.40 & 83.21\\
    GaitBase  & 32    &  141.96  & 60.12  & 75.47 \\
    \midrule[0.6pt]
    Dorefa-4  & w4/a4  & -  & 53.31 & 69.68 \\
    PACT-4 & w4/a4  & -  & 52.40  & 68.72 \\
    LSQ-4 & w4/a4  & -  & 57.44  & 73.51 \\
    QGait-4 & w4/a4  & 2.22 & \textbf{58.52}  & \textbf{74.06} \\
    \midrule[0.6pt]
    Dorefa-8& w8/a8  & -  & 60.00  & 74.93  \\
    PACT-8  & w8/a8  & -  & 59.03  & 73.62 \\
    LSQ-8 & w8/a8  & -  & 59.89  & 75.13  \\
    QGait-8 & w8/a8  & 8.87  & \textbf{60.60} & \textbf{75.89}\\
    \bottomrule[1.2pt]
    \end{tabular}
    \label{tab:2}%
\end{table}%
    
\subsection{Experimental Settings}
\label{sec:4}
\paragraph{Training Settings}
We pre-process the original gait sequences for all the datasets following previous works~\cite{Fan2022OpenGaitRG, Wang2023DyGaitED, Fan2023ExploringDM} and the size of each frame is set to 64 × 44. For Gait3D~\cite{Zheng2022GaitRI}, GREW~\cite{Zhu2021GaitRI} and OUMVLP~\cite{Takemura2018MultiviewLP}, the number of Identities and the number of samples per identity is set to $32 \times 4$ and $8 \times 16$ for CASIA-B~\cite{Yu2006AFF}. The optimizer is Adam with a learning rate of 1e-4 and momentum of 0.9 for all the datasets for a fair comparison. The frame number settings are consistent with OpenGait~\cite{Fan2022OpenGaitRG}. In the first training stage of QGait, the iterations are set to 60K, 180K, and 120Krespectively for Gait3D~\cite{Zheng2022GaitRI}, GREW~\cite{Zhu2021GaitRI} and OUMVLP~\cite{Takemura2018MultiviewLP}. In the fine-tuning stage, we only conduct 1K or 2K iterations according to different datasets to avoid overfitting while ensuring a fair comparison relative to the full-precision model.

\paragraph{Evaluation Settings}
We conducted the evaluation on the test set corresponding to the training dataset. In addition to comparing with quantized models, we also included comparisons with state-of-the-art full-precision models, including GaitSet~\cite{Chao2018GaitSetRG}, GaitPart~\cite{Fan2020GaitPartTP}, GaitGL~\cite{Lin2020GaitRV}, DyGait~\cite{Wagg2004OnAM}  and GaitBase~\cite{Fan2022OpenGaitRG}. Evaluation metrics used include Rank-n, mAP (mean Average Precision), mINP (mean Inverted Normalized Precision), and BitOPs (Bit Operations Per Second). Conventionally, given the weight $\mathbf {w} \in \mathbb{R}^{C \times C_{out} \times F \times F}$ of $b_w$ -bit and input feature $\mathbf {x} \in \mathbb{R}^{N \times C_{in} \times H \times W}$ of $b_a$ -bit, BitOPs of a quantized convolution layer can be calculated as $\frac{b_w}{32}\cdot\frac {b_a}{32}\cdot2C_{in}C_{out}F^2NHW$. The BitOPs of each model on different datasets are calculated based on different frames (100 frames for Gait3D and OUMVLP, 120 frames for GREW), along with different model configurations. 

    \begin{table}[!t]
      \centering
        \caption{Results on OUMVLP dataset.}
  \begin{tabular}{c|cccc}
    \toprule[1.2pt]
    Methods & Bit  & BitOPs & Rank-1 & Rank-5 \\
    \midrule[0.6pt]
    GaitSet  & 32  & 42.29 & 87.12 & 91.64  \\
    GaitPart & 32   & 26.44  & 88.68  & 92.07\\
    GaitGL  & 32  &  193.30  & 89.73 & 92.32 \\
    DyGait & 32  & 829.79 & 90.47 & 92.89\\
    GaitBase  & 32  & 118.30    & 90.32  & 92.67 \\
    \midrule[0.6pt]
    Dorefa-4  & w4/a4 & - & 88.24  & 92.02 \\
    PACT-4 & w4/a4 &- & 87.65  & 91.96 \\
    LSQ-4  & w4/a4 & - & 86.19  & 91.19  \\
    QGait-4 & w4/a4 & 1.85 & \textbf{89.24}  & \textbf{92.29} \\
    \midrule[0.6pt]
    Dorefa-8  & w8/a8 & - & 90.26  & 92.60 \\
    PACT-8 & w8/a8 & -& 90.13  & 92.50 \\
    LSQ-8 & w8/a8 &- & 90.28  & 92.61 \\
    QGait-8 & w8/a8 & 7.39 & \textbf{90.35}  & \textbf{92.63}  \\
    \bottomrule[1.2pt]
    \end{tabular}
    \label{tab:3}%
\end{table}%

\subsection{Comparison with State-of-the-art Methods}
\label{sec:4.2}

Figure~\ref{tab:1}, Fig.~\ref{tab:2}, and Fig.~\ref{tab:3} respectively provide comparisons between QGait and various SOTA models on different datasets. QGait achieves the optimal balance between model performance and computational complexity. For complex datasets like Gait3D, the introduction of quantization may act as a form of regularization, enhancing the network's generalization ability.

\subsection{Ablation Study}

\begin{table}[t]
  \centering
      \caption{Rank-1 accuracy on Gait3D dataset are reported. Ratio means increase of k / iterations.}
    \begin{tabular}{c|cc|cc|c}
    \toprule[1.2pt]
    Strategy & T=3 & ratio & T=5 & ratio & STE\\
    \midrule[0.48pt]
    FIXED &  62.72    & -  & 62.02 & - &  64.80  \\
    GROW-1 & 64.88    & 0.1/100  & 65.48 & 0.1/100 &  - \\
    GROW-2 & 65.24    & 0.2/100  & \textbf{65.89} & 0.2/100 &  -  \\
    GROW-3 & 61.15    & 1/1000  & 58.92 & 1/1000 &  -  \\
    \bottomrule[1.2pt]
    \end{tabular}
    \label{tab:4}
\end{table}%
\begin{table}[t]
  \centering
  \caption{FT denotes the finetuning stage.}
    \begin{tabular}{cccc|c}
    \toprule[1.2pt]
    Baseline & Vanilla KD & IDC & FT & Rank-1 (\%) \\
    \midrule[0.48pt]
    \checkmark & & & & 64.80  \\
    \checkmark & \checkmark & & & 64.21\\
    \checkmark & & \checkmark & & 65.64 \\
    \checkmark & & \checkmark &  \checkmark & \textbf{66.50} \\
    \bottomrule[1.2pt]
    \end{tabular}
    
    \label{tab:5}
\end{table}%

\begin{table}[t]
  \centering
      \caption{Results of different models after quantization on CASIA-B dataset.}
  \begin{tabular}{c|ccc}
    \toprule[1.2pt]
    Methods & R1-NM & R1-BG & R1-CL \\
    \midrule[0.6pt]
    GaitSet~\cite{Chao2018GaitSetRG} & 95.79 & 89.58 & 73.62  \\
    GaitSet-4bit & 95.12   & 88.65 & 72.33 \\
    GaitSet-8bit & 95.73   & 89.44 & 73.58 \\
    \midrule[0.6pt]
    GaitPart~\cite{Fan2020GaitPartTP} & 96.14   & 90.69 & 78.73\\
    GaitPart-4bit & 95.89   & 89.99 & 78.01 \\
    GaitPart-8bit& 96.15   & 90.77 & 78.92 \\
    \midrule[0.6pt]
    DyGait~\cite{Wang2023DyGaitED} & 98.52  & 96.13  & 87.71 \\
    DyGait-4bit & 98.72  & 96.65  & 87.90  \\
    DyGait-8bit & 98.53  & 96.12  & 87.73 \\
    \bottomrule[1.2pt]
    \end{tabular}
    \label{tab:6}
\end{table}

\begin{figure}[!t]
    \centering
    \includegraphics[width=7.5cm]{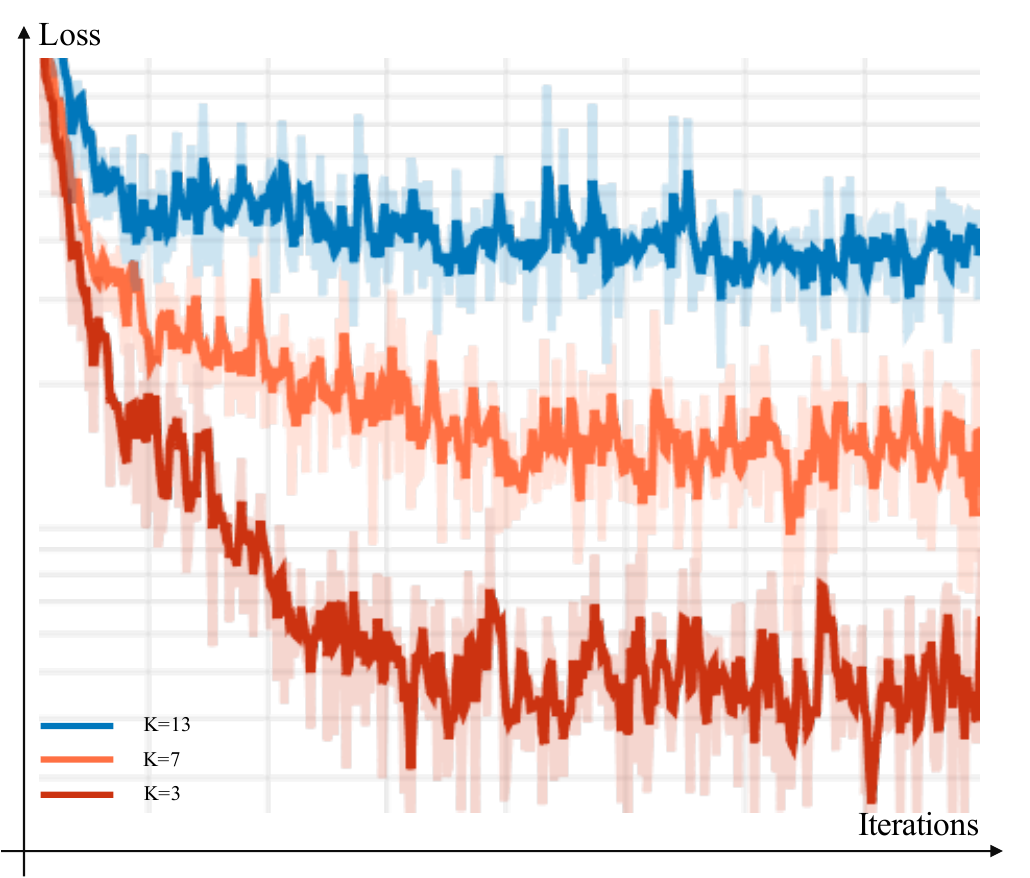}
    \caption{Loss curves under different $k$.}
    \label{fig:kloss}
\end{figure}

\paragraph{Training with Soft Quantizer}
We empirically demonstrate that introducing a soft quantizer at the beginning of training indeed significantly affects the model's convergence. As the parameter $k$ increases, the soft quantizer approaches the round function more closely, resulting in smaller updates to gradients at each step as Fig.~\ref{fig:kloss} shows. We also experimentally determine how to update the $k$ value during the fine-tuning stage. After trying several different strategies, we decided to use a gradually increasing approach, where we increment the value of k as the number of iterations increases until it reaches a predetermined threshold $T$ as Fig.~\ref{tab:4} shows. 

\paragraph{Inter-class Distance-guided Calibration} 
To study the effects of IDC together with other components, we conducted module ablation experiments on the baseline method LSQ. It can be observed from Fig.~\ref{tab:5} that, as discussed in the methodology section, vanilla KD overly balances the quantization error, resulting in decreased accuracy. IDC improves the accuracy by 1.84\%, and even after incorporating the soft quantizer fine-tuning strategy, it still achieves a further 0.86\% improvement in accuracy.

\subsection{Quantization of Different Models}
\label{sec:4.4}
We also conduct experiments on different model architectures using our proposed QGait method, including GaitSet, GaitPart and DyGait. It can be seen from Fig.~\ref{tab:6} that QGait shows comparable performance with full-precision models on different model structures. 
For two lightweight models, the 4-bit quantized models lost some precision. However, the quantized 4-bit DyGait model outperforms the full-precision model. That is because the quantization acts as a regularization. It can alleviate overfitting on some heavy models while the data is relatively simple. This property may inspire the later works.

\subsection{Quantization of Different Backbones}
Future gait models may require larger-scale backbones, rather than being limited to just ResNet9. To demonstrate the effectiveness of quantization, we individually quantized different backbones to observe their latency as Fig.~\ref{tab:qbackbone} shows.

\begin{table}[!t]
  \centering
    \caption{Quantization of Different Backbones on Tesla T4. The batch size is set to 32. Average per-sample latency is reported.}
    \begin{tabular}{c|cccc}
    \toprule[1.2pt]
     & Bit & Params & BitOPs & Latency  \\
    \midrule[0.6pt]
    ResNet-18  & 32    & 44.6M  & 1858G & 18.2ms  \\
    ResNet-18 & 8    & 11.1M  & 116G  & 5.7ms \\
    ResNet-18  & 4    & 5.8M  & 34G & 4.1ms \\
    ResNet-50  & 32    &  97.8M  & 6951G & 50.6ms\\
    ResNet-50  & 8    &  24.5M  & 247G  & 15.9ms \\
    ResNet-50  & 4    &  13.1M  & 67G  & 11.4ms \\
    \bottomrule[1.2pt]
    \end{tabular}
    \label{tab:qbackbone}%
\end{table}%

\section{Conclusion}
In this paper, we explore a gait representation network based on quantization compression for the first time. Addressing the data characteristics of gait representation learning tasks and the difficulty of training convergence, we design quantization schemes that are different from conventional tasks. Experimental results demonstrate that the two proposed methods effectively improve the quantization accuracy of gait representation learning. In experiments, we found that different datasets exhibit varying levels of robustness to quantization, which may be related to the complexity of the data. This brings new perspectives and considerations for future research in gait representation learning.

{\small
\bibliographystyle{ieee}
\bibliography{egbib}

@article{Connor2018BiometricRB,
  title={Biometric recognition by gait: A survey of modalities and features},
  author={Patrick Connor and Arun Ross},
  journal={Comput. Vis. Image Underst.},
  year={2018},
  volume={167},
  pages={1-27}
}

@article{Lam2007HumanGR,
  title={Human gait recognition by the fusion of motion and static spatio-temporal templates},
  author={Toby H. W. Lam and Raymond S. H. Lee and Dafan Zhang},
  journal={Pattern Recognit.},
  year={2007},
  volume={40},
  pages={2563-2573}
}

@article{Guan2015OnRT,
  title={On Reducing the Effect of Covariate Factors in Gait Recognition: A Classifier Ensemble Method},
  author={Yu Guan and Chang-Tsun Li and Fabio Roli},
  journal={IEEE Transactions on Pattern Analysis and Machine Intelligence},
  year={2015},
  volume={37},
  pages={1521-1528}
}

@article{Choudhury2016ClothingAC,
  title={Clothing and carrying condition invariant gait recognition based on rotation forest},
  author={Sruti Das Choudhury and Tardi Tjahjadi},
  journal={Pattern Recognit. Lett.},
  year={2016},
  volume={80},
  pages={1-7}
}

@article{Wagg2004OnAM,
  title={On automated model-based extraction and analysis of gait},
  author={David Kenneth Wagg and Mark S. Nixon},
  journal={Sixth IEEE International Conference on Automatic Face and Gesture Recognition, 2004. Proceedings.},
  year={2004},
  pages={11-16}
}

@article{Wu2017ACS,
  title={A Comprehensive Study on Cross-View Gait Based Human Identification with Deep CNNs},
  author={Zifeng Wu and Yongzhen Huang and Liang Wang and Xiaogang Wang and Tieniu Tan},
  journal={IEEE Transactions on Pattern Analysis and Machine Intelligence},
  year={2017},
  volume={39},
  pages={209-226}
}

@article{Makihara2017JointIA,
  title={Joint Intensity and Spatial Metric Learning for Robust Gait Recognition},
  author={Yasushi Makihara and Atsuyuki Suzuki and Daigo Muramatsu and Xiang Li and Yasushi Yagi},
  journal={2017 IEEE Conference on Computer Vision and Pattern Recognition (CVPR)},
  year={2017},
  pages={6786-6796}
}

@article{Fan2020GaitPartTP,
  title={GaitPart: Temporal Part-Based Model for Gait Recognition},
  author={Chao Fan and Yunjie Peng and Chunshui Cao and Xu Liu and Saihui Hou and Jiannan Chi and Yongzhen Huang and Qing Li and Zhiqiang He},
  journal={2020 IEEE/CVF Conference on Computer Vision and Pattern Recognition (CVPR)},
  year={2020},
  pages={14213-14221}
}

@inproceedings{Hou2020GaitLN,
  title={Gait Lateral Network: Learning Discriminative and Compact Representations for Gait Recognition},
  author={Saihui Hou and Chunshui Cao and Xu Liu and Yongzhen Huang},
  booktitle={European Conference on Computer Vision},
  year={2020}
}

@article{Yu2006AFF,
  title={A Framework for Evaluating the Effect of View Angle, Clothing and Carrying Condition on Gait Recognition},
  author={Shiqi Yu and Daoliang Tan and Tieniu Tan},
  journal={18th International Conference on Pattern Recognition (ICPR'06)},
  year={2006},
  volume={4},
  pages={441-444}
}

@article{Takemura2018MultiviewLP,
  title={Multi-view large population gait dataset and its performance evaluation for cross-view gait recognition},
  author={Noriko Takemura and Yasushi Makihara and Daigo Muramatsu and Tomio Echigo and Yasushi Yagi},
  journal={IPSJ Transactions on Computer Vision and Applications},
  year={2018},
  volume={10},
  pages={1-14}
}

@article{Fan2022OpenGaitRG,
  title={OpenGait: Revisiting Gait Recognition Toward Better Practicality},
  author={Chao Fan and Junhao Liang and Chuanfu Shen and Saihui Hou and Yongzhen Huang and Shiqi Yu},
  journal={2023 IEEE/CVF Conference on Computer Vision and Pattern Recognition (CVPR)},
  year={2022},
  pages={9707-9716}
}

@misc{Fan2023ExploringDM,
  title={Exploring Deep Models for Practical Gait Recognition},
  author={Chao Fan and Saihui Hou and Yongzhen Huang and Shiqi Yu},
  journal={ArXiv},
  year={2023},
  volume={abs/2303.03301},
  eprint="2303.03301",
  archivePrefix="arxiv"
}

@article{Zheng2022GaitRI,
  title={Gait Recognition in the Wild with Dense 3D Representations and A Benchmark},
  author={Jinkai Zheng and Xinchen Liu and Wu Liu and Lingxiao He and Chenggang Clarence Yan and Tao Mei},
  journal={2022 IEEE/CVF Conference on Computer Vision and Pattern Recognition (CVPR)},
  year={2022},
  pages={20196-20205}
}

@article{Lin2020GaitRV,
  title={Gait Recognition via Effective Global-Local Feature Representation and Local Temporal Aggregation},
  author={Beibei Lin and Shunli Zhang and Xin Yu},
  journal={2021 IEEE/CVF International Conference on Computer Vision (ICCV)},
  year={2020},
  pages={14628-14636}
}

@article{Shen2022LidarGaitB3,
  title={LidarGait: Benchmarking 3D Gait Recognition with Point Clouds},
  author={Chuanfu Shen and Chao Fan and Wei Wu and Rui Wang and George Q. Huang and Shiqi Yu},
  journal={2023 IEEE/CVF Conference on Computer Vision and Pattern Recognition (CVPR)},
  year={2022},
  pages={1054-1063}
}

@article{He2018FilterPV,
  title={Filter Pruning via Geometric Median for Deep Convolutional Neural Networks Acceleration},
  author={Yang He and Ping Liu and Ziwei Wang and Zhilan Hu and Yi Yang},
  journal={2019 IEEE/CVF Conference on Computer Vision and Pattern Recognition (CVPR)},
  year={2018},
  pages={4335-4344}
}

@article{Lin2020HRankFP,
  title={HRank: Filter Pruning Using High-Rank Feature Map},
  author={Mingbao Lin and Rongrong Ji and Yan Wang and Yichen Zhang and Baochang Zhang and Yonghong Tian and Ling Shao},
  journal={2020 IEEE/CVF Conference on Computer Vision and Pattern Recognition (CVPR)},
  year={2020},
  pages={1526-1535}
}

@misc{Zhou2021LearningNM,
  title={Learning N: M Fine-grained Structured Sparse Neural Networks From Scratch},
  author={Aojun Zhou and Yukun Ma and Junnan Zhu and Jianbo Liu and Zhijie Zhang and Kun Yuan and Wenxiu Sun and Hongsheng Li},
  journal={ArXiv},
  year={2021},
  volume={abs/2102.04010},
  eprint="2102.04010",
  archivePrefix="arxiv"
}

@article{Lee2021NetworkQW,
  title={Network Quantization with Element-wise Gradient Scaling},
  author={Junghyup Lee and Dohyung Kim and Bumsub Ham},
  journal={2021 IEEE/CVF Conference on Computer Vision and Pattern Recognition (CVPR)},
  year={2021},
  pages={6444-6453}
}

@misc{Esser2019LearnedSS,
  title={Learned Step Size Quantization},
  author={Steven K. Esser and Jeffrey L. McKinstry and Deepika Bablani and Rathinakumar Appuswamy and Dharmendra S. Modha},
  journal={ArXiv},
  year={2019},
  volume={abs/1902.08153},
  eprint="1902.08153",
  archivePrefix="arxiv"
}

@article{Choi2018PACTPC,
  title={PACT: Parameterized Clipping Activation for Quantized Neural Networks},
  author={Jungwook Choi and Zhuo Wang and Swagath Venkataramani and Pierce I-Jen Chuang and Vijayalakshmi Srinivasan and K. Gopalakrishnan},
  journal={ArXiv},
  year={2018},
  volume={abs/1805.06085}
}

@misc{Zhou2016DoReFaNetTL,
  title={DoReFa-Net: Training Low Bitwidth Convolutional Neural Networks with Low Bitwidth Gradients},
  author={Shuchang Zhou and Zekun Ni and Xinyu Zhou and He Wen and Yuxin Wu and Yuheng Zou},
  journal={ArXiv},
  year={2016},
  volume={abs/1606.06160},
  eprint="1606.06160",
  archivePrefix="arxiv"
}

@article{Jacob2017QuantizationAT,
  title={Quantization and Training of Neural Networks for Efficient Integer-Arithmetic-Only Inference},
  author={Benoit Jacob and Skirmantas Kligys and Bo Chen and Menglong Zhu and Matthew Tang and Andrew G. Howard and Hartwig Adam and Dmitry Kalenichenko},
  journal={2018 IEEE/CVF Conference on Computer Vision and Pattern Recognition},
  year={2017},
  pages={2704-2713}
}

@inproceedings{Zhang2018LQNetsLQ,
  title={LQ-Nets: Learned Quantization for Highly Accurate and Compact Deep Neural Networks},
  author={Dongqing Zhang and Jiaolong Yang and Dongqiangzi Ye and Gang Hua},
  booktitle={European Conference on Computer Vision},
  year={2018}
}

@misc{Baskin2018NICENI,
  title={NICE: Noise Injection and Clamping Estimation for Neural Network Quantization},
  author={Chaim Baskin and Natan Liss and Yoav Chai and Evgenii Zheltonozhskii and Eli Schwartz and Raja Giryes and Avi Mendelson and Alexander M. Bronstein},
  journal={ArXiv},
  year={2018},
  volume={abs/1810.00162},
  eprint="1810.00162",
  archivePrefix="arxiv"
}

@misc{Yao2022ZeroQuantEA,
  title={ZeroQuant: Efficient and Affordable Post-Training Quantization for Large-Scale Transformers},
  author={Zhewei Yao and Reza Yazdani Aminabadi and Minjia Zhang and Xiaoxia Wu and Conglong Li and Yuxiong He},
  journal={ArXiv},
  year={2022},
  volume={abs/2206.01861},
  eprint="2206.01861",
  archivePrefix="arxiv"
}

@misc{Kim2019QKDQK,
  title={QKD: Quantization-aware Knowledge Distillation},
  author={Jangho Kim and Yash Bhalgat and Jinwon Lee and Chirag Patel and Nojun Kwak},
  journal={ArXiv},
  year={2019},
  volume={abs/1911.12491},
  eprint="1911.12491",
  archivePrefix="arxiv"
}

@inproceedings{Li2020EndtoEndMG,
  title={End-to-End Model-Based Gait Recognition},
  author={Xiang Li and Yasushi Makihara and Chi Xu and Yasushi Yagi and Shiqi Yu and Mingwu Ren},
  booktitle={Asian Conference on Computer Vision},
  year={2020}
}

@article{Wang2023DyGaitED,
  title={DyGait: Exploiting Dynamic Representations for High-performance Gait Recognition},
  author={Ming-Zhen Wang and Xianda Guo and Beibei Lin and Tian Yang and Zhenguo Zhu and Lincheng Li and Shunli Zhang and Xin Yu},
  journal={2023 IEEE/CVF International Conference on Computer Vision (ICCV)},
  year={2023},
  pages={13378-13387}
}

@inproceedings{Liao2017PoseBasedTN,
  title={Pose-Based Temporal-Spatial Network (PTSN) for Gait Recognition with Carrying and Clothing Variations},
  author={Rijun Liao and Chunshui Cao and Edel B. Garc{\'i}a Reyes and Shiqi Yu and Yongzhen Huang},
  booktitle={Chinese Conference on Biometric Recognition},
  year={2017}
}

@article{Shen2022GaitRW,
  title={Gait Recognition with Mask-based Regularization},
  author={Chuanfu Shen and Beibei Lin and Shunli Zhang and George Q. Huang and Shiqi Yu and Xin-cen Yu},
  journal={2023 IEEE International Joint Conference on Biometrics (IJCB)},
  year={2022},
  pages={1-10}
}

@article{Loper2023SMPLAS,
  title={SMPL: A Skinned Multi-Person Linear Model},
  author={Matthew Loper and Naureen Mahmood and Javier Romero and Gerard Pons-Moll and Michael J. Black},
  journal={Seminal Graphics Papers: Pushing the Boundaries, Volume 2},
  year={2023}
}

@article{Teepe2021GaitgraphGC,
  title={Gaitgraph: Graph Convolutional Network for Skeleton-Based Gait Recognition},
  author={Torben Teepe and Ali R. Khan and Johannes Gilg and Fabian Herzog and Stefan H{\"o}rmann and Gerhard Rigoll},
  journal={2021 IEEE International Conference on Image Processing (ICIP)},
  year={2021},
  pages={2314-2318}
}

@article{Zhang2022SpatialTN,
  title={Spatial transformer network on skeleton‐based gait recognition},
  author={Cun Zhang and Xingyun Chen and Guohui Han and Xiangrong Liu},
  journal={Expert Systems},
  year={2022},
  volume={40}
}

@inproceedings{Chao2018GaitSetRG,
  title={GaitSet: Regarding Gait as a Set for Cross-View Gait Recognition},
  author={Hanqing Chao and Yiwei He and Junping Zhang and Jianfeng Feng},
  booktitle={AAAI Conference on Artificial Intelligence},
  year={2018}
}

@misc{Chen2024AdaptiveQW,
  title={Adaptive quantization with mixed-precision based on low-cost proxy},
  author={Jing Chen and Qiao Yang and Senmao Tian and Shunli Zhang},
  journal={ArXiv},
  year={2024},
  volume={abs/2402.17706},
  eprint="2402.17706",
  archivePrefix="arxiv"
}

@misc{Rastegari2016XNORNetIC,
  title={XNOR-Net: ImageNet Classification Using Binary Convolutional Neural Networks},
  author={Mohammad Rastegari and Vicente Ordonez and Joseph Redmon and Ali Farhadi},
  journal={ArXiv},
  year={2016},
  volume={abs/1603.05279},
  eprint="1603.05279",
  archivePrefix="arxiv"
}

@misc{Cai2023BinarizedSC,
  title={Binarized Spectral Compressive Imaging},
  author={Yuan-Yuan Cai and Yuxing Zheng and Jing Lin and Haoqian Wang and Xin Yuan and Yulun Zhang},
  journal={ArXiv},
  year={2023},
  volume={abs/2305.10299},
  eprint="2305.10299",
  archivePrefix="arxiv"
}

@article{Liu2021NonuniformtoUniformQT,
  title={Nonuniform-to-Uniform Quantization: Towards Accurate Quantization via Generalized Straight-Through Estimation},
  author={Zechun Liu and Kwang-Ting Cheng and Dong Huang and Eric P. Xing and Zhiqiang Shen},
  journal={2022 IEEE/CVF Conference on Computer Vision and Pattern Recognition (CVPR)},
  year={2021},
  pages={4932-4942}
}

@article{Shang2022PostTrainingQO,
  title={Post-Training Quantization on Diffusion Models},
  author={Yuzhang Shang and Zhihang Yuan and Bin Xie and Bingzhe Wu and Yan Yan},
  journal={2023 IEEE/CVF Conference on Computer Vision and Pattern Recognition (CVPR)},
  year={2022},
  pages={1972-1981}
}

@article{Tian2023CABMCB,
  title={CABM: Content-Aware Bit Mapping for Single Image Super-Resolution Network with Large Input},
  author={Senmao Tian and Ming Lu and Jiaming Liu and Yandong Guo and Yurong Chen and Shunli Zhang},
  journal={2023 IEEE/CVF Conference on Computer Vision and Pattern Recognition (CVPR)},
  year={2023},
  pages={1756-1765}
}

@misc{Lin2023AWQAW,
  title={AWQ: Activation-aware Weight Quantization for LLM Compression and Acceleration},
  author={Ji Lin and Jiaming Tang and Haotian Tang and Shang Yang and Xingyu Dang and Song Han},
  journal={ArXiv},
  year={2023},
  volume={abs/2306.00978},
  eprint="2306.00978",
  archivePrefix="arxiv"
}

@misc{Bengio2013EstimatingOP,
  title={Estimating or Propagating Gradients Through Stochastic Neurons for Conditional Computation},
  author={Yoshua Bengio and Nicholas L{\'e}onard and Aaron C. Courville},
  journal={ArXiv},
  year={2013},
  volume={abs/1308.3432},
  eprint="1308.3432",
  archivePrefix="arxiv"
}

@article{Schroff2015FaceNetAU,
  title={FaceNet: A unified embedding for face recognition and clustering},
  author={Florian Schroff and Dmitry Kalenichenko and James Philbin},
  journal={2015 IEEE Conference on Computer Vision and Pattern Recognition (CVPR)},
  year={2015},
  pages={815-823}
}

@article{Zhu2021GaitRI,
  title={Gait Recognition in the Wild: A Benchmark},
  author={Zheng Zhu and Xianda Guo and Tian Yang and Junjie Huang and Jiankang Deng and Guan Huang and Dalong Du and Jiwen Lu and Jie Zhou},
  journal={2021 IEEE/CVF International Conference on Computer Vision (ICCV)},
  year={2021},
  pages={14769-14779}
}

@article{Lin2020GaitRW,
  title={Gait Recognition with Multiple-Temporal-Scale 3D Convolutional Neural Network},
  author={Beibei Lin and Shunli Zhang and Feng Bao},
  journal={Proceedings of the 28th ACM International Conference on Multimedia},
  year={2020},
  url={https://api.semanticscholar.org/CorpusID:222278506}
}

@article{guo2025gait,
  title={Gait recognition in the wild: A large-scale benchmark and NAS-based baseline},
  author={Guo, Xianda and Zhu, Zheng and Yang, Tian and Lin, Beibei and Huang, Junjie and Deng, Jiankang and Huang, Guan and Zhou, Jie and Lu, Jiwen},
  journal={IEEE Transactions on Pattern Analysis and Machine Intelligence},
  year={2025},
  publisher={IEEE}
}

@article{wang2025gaitadapt,
  title={GaitAdapt: Continual Learning for Evolving Gait Recognition},
  author={Wang, Jingjie and Zhang, Shunli and Wei, Xiang and Tian, Senmao},
  journal={arXiv preprint arXiv:2508.03375},
  year={2025}
}
}

\end{document}